\def\year{2017}\relax
\documentclass[letterpaper]{article}
\usepackage{aaai17}
\usepackage{times}
\usepackage{helvet}
\usepackage{courier}
\usepackage{amsmath}
\usepackage{amssymb}
\usepackage{amsopn}
\usepackage{times}
\usepackage{latexsym}
\usepackage{subcaption}
\usepackage{CJKutf8}
\usepackage{graphicx}
\usepackage{multirow}
\usepackage[english]{babel}
\usepackage{bm}

\usepackage{CJK}
\AtBeginDvi{\input{zhwinfonts}}
\begin{CJK*}{GBK}{kai}

\begin{document}
%
\title{Lattice-Based Recurrent Neural Network Encoders for \\ Neural Machine Translation}
\author{Jinsong Su$^{1}$, \ \ Zhixing Tan$^{1}$,  \ \ Deyi Xiong$^{2}$,  \ \ Rongrong Ji$^{1}$,  \ \ Xiaodong Shi$^{1}$,  \ \ Yang Liu$^{3}\thanks{\ \ Corresponding author.}$\\
  Xiamen University, Xiamen, China$^{1}$ \\
  Soochow University, Suzhou, China$^{2}$ \\
  Tsinghua University, Beijing, China$^{3}$ \\
  {\tt \{jssu,\ rrji,\ mandel\}@xmu.edu.cn, playinf@stu.xmu.edu.cn} \\
  {\tt dyxiong@suda.edu.cn, liuyang2011@tsinghua.edu.cn} \\
  }
\maketitle

\begin{abstract}
Neural machine translation (NMT) heavily relies on word-level modelling to learn semantic representations of input sentences.
However, for languages without natural word delimiters (e.g., Chinese) where input sentences have to be tokenized first,
conventional NMT is confronted with two issues:
1) it is difficult to find an optimal tokenization granularity for source sentence modelling, and
2) errors in 1-best tokenizations may propagate to the encoder of NMT.
To handle these issues, we propose word-lattice based Recurrent Neural Network (RNN) encoders for NMT,
which generalize the standard RNN to word lattice topology.
The proposed encoders take as input a word lattice that compactly encodes multiple tokenizations, and learn to generate new hidden states from arbitrarily many inputs and hidden states in preceding time steps.
As such, the word-lattice based encoders not only alleviate the negative impact of tokenization errors but also are more expressive and flexible to embed input sentences.
Experiment results on Chinese-English translation demonstrate the superiorities of the proposed encoders over the conventional encoder.
\end{abstract}

\section{Introduction}
In the last two years, 
NMT \cite{Kalchbrenner:EMNLP13,Sutskever:NIPS14,Bahdanau:ICLR15} has obtained state-of-the-art translation performance on some language pairs.
In contrast to conventional statistical machine translation that models latent structures and correspondences between the source and target languages in a pipeline,
NMT trains a unified encoder-decoder \cite{Cho:EMNLP14,Sutskever:NIPS14} neural network for translation,
where an encoder maps the input sentence into a fixed-length vector, and a decoder generates a translation from the encoded vector.

Most NMT systems adopt RNNs 
such as Long Short-Term Memory (LSTM) \cite{Hochreiter:NC97} and Gated Recurrent Unit (GRU) \cite{Cho:EMNLP14} to learn vector representations of source sentences and to generate target sentences due to the strong capability of RNNs in capturing long-distance dependencies.
Generally, such representations and generations are learned and performed at the word level.
Very recently, we have also seen successful approaches in character-level NMT \cite{Wang:Arxiv15,Marta:ACL16,Chung:16}.
In these methods, each input sentence is first segmented into a sequence of words, after which RNNs are used to learn word representations and perform generations at the character level.
Note that all approaches so far exploit word boundary information for learning source representations.
Such a modeling philosophy works well for languages like English.
However, it is not a desirable choice for languages without natural word delimiters such as Chinese.
The underlying reasons are twofold:
1) the optimal tokenization granularity is always a paradox for machine translation since coarse granularity causes data sparseness while fine granularity results in the loss of useful information,
and 2) there may exist some errors in 1-best tokenizations, which potentially propagate to source sentence representations.
Therefore, it is necessary to offer multiple tokenizations instead of a single tokenized sequence to NMT for better source sentence modelling.

In order to handle these tokenization issues, we propose word-lattice based RNN encoders for NMT.
Word lattice is a packed representation of many tokenizations, which has been successfully used in a variety of NLP tasks \cite{Xu:IWSLT05,Dyer:08,Jiang:08,Wang:13}.
Generalizing the standard RNN to word lattice topology,
we expect to not only eliminate the negative effect caused by 1-best tokenization errors but also make the encoder more expressive and flexible to learn semantic representations of source sentences.
In this work, we investigate and compare two word-lattice based RNN encoders:
1) a \emph{Shallow Word-Lattice Based GRU Encoder} which builds on the combinations of inputs and hidden states derived from multiple tokenizations without any change to the architecture of the standard GRU,
and 2) a \emph{Deep Word-Lattice Based GRU Encoder} which learns and updates tokenization-specific vectors for gates, inputs and hidden states, and then generates hidden state vectors for current units.
In both encoders, many different tokenizations can be simultaneously exploited for input sentence modeling.

To demonstrate the effectiveness of the proposed encoders, we carry out experiments on Chinese-English translation tasks.
Experiment results show that: (1) it is really necessary to exploit word boundary information for learning accurate embeddings of input Chinese sentences; (2) word-lattice based RNN encoders outperform the standard RNN encoder in NMT.
To the best of our knowledge, this is the first attempt to build NMT on word lattices.

\section{Neural Machine Translation}
The dominant NMT model is an attention-based one
\cite{Bahdanau:ICLR15}, which includes an encoder network and a decoder network with attention mechanism.

The encoder is generally a bidirectional RNN.
The forward RNN reads a source sentence $\mathbf{x}$=$x_1,x_2...x_I$ from left to right and applies the recurrent activation function $\phi$ to learn semantic representation of word sequence $x_{1:i}$ as $\overrightarrow{h}_i$=$\phi(\overrightarrow{h}_{i-1}, \vec{x}_i)$.
Similarly, the backward RNN scans the source sentence in a reverse direction and learns the semantic representation $\overleftarrow{h}_i$ of the word sequence $x_{i:I}$.
Finally, we concatenate the hidden states of the two RNNs to form the source annotation $h_i=[\overrightarrow{h}^T_i,\overleftarrow{h}^T_i]^T$,
which encodes information about the $i$-th word with respect to all the other surrounding words in the source sentence.

The decoder is a forward RNN producing the translation $y$ in the following way:
\begin{equation} \label{produce_translation}
p(y_j|y_{<j}, \mathbf{x}) = g(y_{j-1}, s_j, m_j),
\end{equation}
here $g(\cdot)$ is a non-linear function, and $s_j$ and $m_j$ denote the decoding state and the source context at the $j$th time step, respectively.
In particular, $s_i$ is computed as follows:
\begin{equation} \label{decoding_state}
s_j = f(s_{j-1}, y_{j-1}, m_j),
\end{equation}
where $f(\cdot)$ is an activation function such as GRU function.
According to the attention mechanism, we define $m_j$ as the weighted sum of the source annotations \{$h_i$\}:
\begin{equation} \label{decoder_hidden}
m_j = \sum\limits_{i=1}^I\alpha_{j,i} \cdot h_i,
\end{equation}
where $\alpha_{j,i}$ measures how well $y_j$ and $h_i$ match as below:
\begin{equation} \label{a_ji}
\alpha_{j,i} = \frac{exp(e_{j,i})}{\sum_{i'=1}^I exp(e_{j,i'})}, \\
\end{equation}
\begin{equation} \label{e_jk}
e_{j,i} = v_a^T tanh(W_as_{j-1} + U_ah_i),
\end{equation}
where $W_a$, $U_a$ and $v_a$ are the weight matrices of attention model.
With this model, the decoder automatically selects words in the source sentence that are relevant to target words being generated.

\section{Word-Lattice based RNN Encoders}
In this section, we study how to incorporate word lattice into the RNN encoder of NMT.
We first briefly review the standard GRU, which is the basis of our encoder units.
Then, we describe how to generate the word lattice of each input sentence.
Finally, we describe word-lattice based GRU encoders in detail.

\subsection{The Standard GRU}
GRU is a new type of hidden unit motivated by LSTM.
As illustrated in Figure \ref{GRU3}(a), at time step $t$, GRU has two gates:
1) a reset gate $r_t$ which determines how to combine the new input $x_t$ with the previous memory $h_{t-1}$,
and 2) an update gate $z_t$ that defines how much of the previous memory to keep around.
Formally, the GRU transition equations are as follows:
\begin{align}
\label{GRU-Input-Beg}
&r_t = \sigma(W^{(r)}x_{t}+U^{(r)}h_{t-1}),\\
&z_t = \sigma(W^{(z)}x_{t}+U^{(z)}h_{t-1}),\\
&\tilde{h}_t = \phi(Wx_{t}+U(r_t\odot h_{t-1})),\\
&h_t = z_t\odot h_{t-1} + (1-z_t)\odot \tilde{h}_t,
\end{align}
where $\sigma$ is the logistic sigmoid function, $\odot$ denotes the elementwise multiplication, and
$W^{*}$ and $U^{*}$ are the weight matrices, respectively.

Similar to LSTM, GRU overcomes \emph{exploding or vanishing gradient problem} \cite{Hochreiter:NC97} of the conventional RNN.
However, GRU is easier to compute and implement than LSTM.
Therefore, in this work, we adopt GRU to build our word lattice based encoders.
However, our method can be applied on other RNNs as well.

\subsection{Word Lattice}
As shown in Figure \ref{Word-Lattice}, a word lattice is a directed graph $G=\langle V, E\rangle$, where $V$ is node set and $E$ is edge set.
Given the word lattice of a character sequence $c_{1:N}$=$c_1...c_N$,
node $v_i\in V$ denotes the position between $c_i$ and $c_{i+1}$,
edge $e_{i:j} \in E$ departs from $v_i$ and arrives at $v_j$ from left to right, covering the subsequence $c_{i+1:j}$ that is recognized as a possible word.
To generate word lattices, we train many word segmenters using multiple corpora with different annotation standards, such as \emph{the Peking University Corpus} (\emph{PKU}), \emph{the Microsoft Research Corpus} (\emph{MSR}) and \emph{the Penn Chinese Treebank 6.0} (\emph{CTB}).
For each input sentence, we tokenize it using these different segmenters and generate a word lattice by merging the predicted tokenizations that are shared in different segmenters.
For example, in Figure \ref{Word-Lattice}, both \emph{CTB} and \emph{MSR} segmenters produce the same tokenization ``副-总-理'', which is merged into the edge $e_{0:3}$ in the lattice.
Obviously, word lattice has richer network topology than a single word sequence.
It encodes many tokenizations for the same character subsequence,
and thus there may exist multiple inputs and preceding hidden states at each time step,
which can not be simultaneously modelled in the standard RNN.

\subsection{Word-Lattice based RNN with GRU Encoders}
\begin{figure}[!t]
\centering
\includegraphics[height=5.6cm,width=9cm]{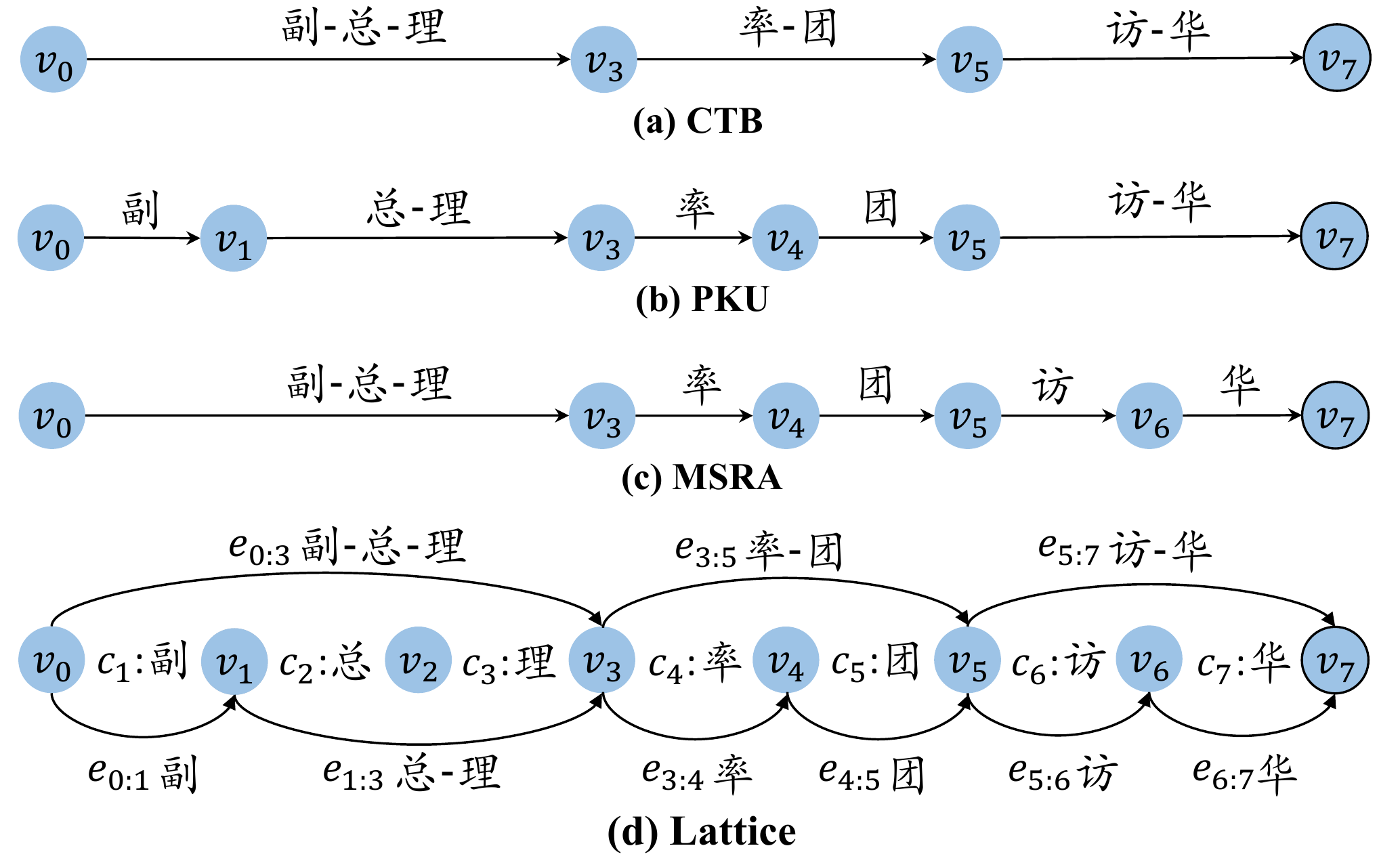}
\caption{\label{Word-Lattice}
Three kinds of word segmentations and the resulting word lattice of the Chinese character sequence ``$c_1$副$c_2$总$c_3$理$c_4$率$c_5$团$c_6$访$c_7$华''.
We insert ``-'' between characters in a word just for clarity.
}
\end{figure}
To deal with the above problem, we propose word-lattice based GRU encoders for NMT.
Similar to the dominant NMT model \cite{Bahdanau:ICLR15}, our encoders are bidirectional RNNs with GRU.
Here we only introduce the forward RNN. The backward RNN can be extended in a similar way.

Our encoders scan a source sentence character by character.
Only at potential word boundaries, hidden states are generated from many input candidate words and the corresponding preceding hidden states.
Specifically, at time step $t$, we first identify all edges pointing to $v_t$, each of which covers different input words with preceding hidden states.
In particular, for the $k$th edge\footnote{We index the edges from left to right according to the positions of their starting nodes.}, we denote its input word vector and the corresponding preceding hidden state as $x_t^{(k)}$ and $h_{pre}^{(k)}$, respectively.
As shown in Figure \ref{Word-Lattice},
the word lattice contains two edges $e_{0:3}$ and $e_{1:3}$, both of which link to $v_3$.
Therefore there exist two input words ``副-总-理'' and ``总-理'' with different preceding hidden states at the $3$rd time step.
We then propose two word-lattice based GRUs to exploit multiple tokenizations simultaneously:

\begin{figure}[!t]
\centering
\includegraphics[height=13cm,width=7.5cm]{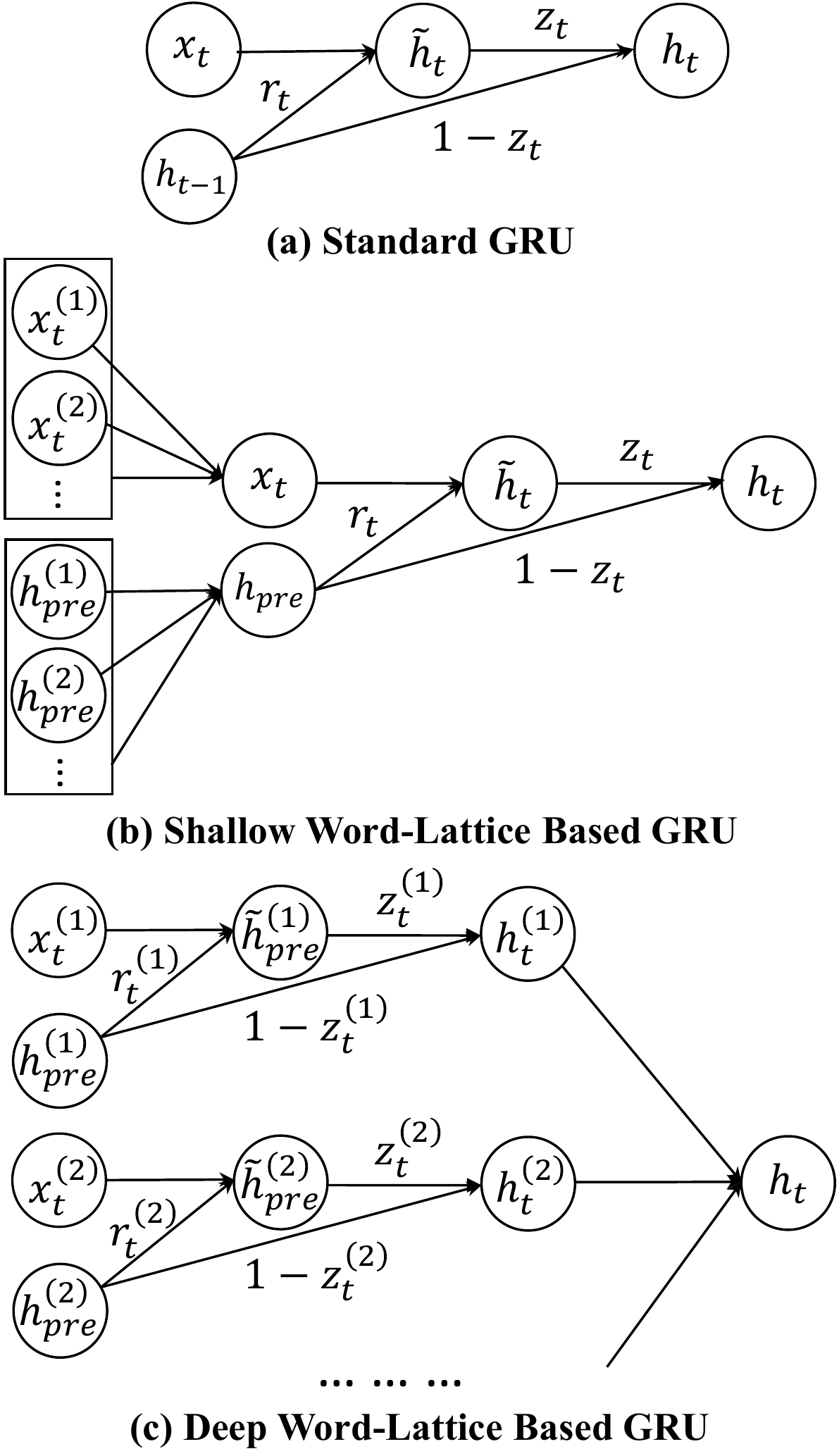}
\caption{
\label{GRU3}
The architectures of the standard GRU and word-lattice based GRUs.
Note that in our GRUs, the preceding hidden state is not always the one at the time step $t$-$1$.
For notational clarity, we use the subscript ``\emph{pre}'' rather than $t$-$1$ to index the preceding hidden state.
}
\end{figure}

(1) \emph{\textbf{Shallow Word-Lattice based GRU}} (\emph{\textbf{SWL-GRU}}).
The architecture of \emph{SWL-GRU} is shown in Figure \ref{GRU3}(b).
At the potential word boundary of the $t$th time step,
we combine all possible word embeddings $\{x_{t}^{*}\}$ into a compressed $x_t$.
Similarly, the hidden state vectors $\{h_{pre}^{*}\}$ of preceding time steps are also compressed into $h_{pre}$.
Then, both $x_t$ and $h_{pre}$ are fed into the standard GRU to generate the final hidden state vector $h_t$.
Obviously, here we do not change the inner architecture of the standard GRU.
The combination of multiple word embeddings into one compressed vector, the combination of all preceding hidden states and the corresponding GRU
are defined as follows:
\begin{align}
\label{Word-Lattice-GRU1-Input-Beg}
&x_{t} = g(x_t^{(1)},x_t^{(2)},...),\\
\label{Word-Lattice-GRU1-Input-End}
&h_{pre} = g(h_{pre}^{(1)},h_{pre}^{(2)},...),\\
&r_t = \sigma(W^{(r)}x_{t}+U^{(r)}h_{pre}),\\
&z_t = \sigma(W^{(z)}x_{t}+U^{(z)}h_{pre}),\\
&\tilde{h}_t = \phi(Wx_{t}+U(r_t\odot h_{pre})),\\
&h_t = z_t\odot h_{t-1} + (1-z_t)\odot \tilde{h}_t,
\end{align}
where Eqs. (\ref{Word-Lattice-GRU1-Input-Beg})-(\ref{Word-Lattice-GRU1-Input-End}) are used to generate the compressed representations $x_t$ and $h_{pre}$,
the others are the same as those of the standard GRU, and $g$(*) is a composition function, for which we will investigate two definitions later on.

(2) \emph{\textbf{Deep Word-Lattice based GRU}} (\emph{\textbf{DWL-GRU}}).
The architecture of \emph{DWL-GRU} is illustrated in Figure \ref{GRU3}(c).
In this unit, we set and update the reset gate $r_{t}^{(k)}$, the update gate $z_{t}^{(k)}$ and the hidden state vector $h_{t}^{(k)}$ that are specific to
the $k$th edge ending with $c_t$,
and then generate a composed hidden state vector $h_t$ from $\{h_{t}^{(*)}\}$.
Different from \emph{SWL-GRU}, \emph{DWL-GRU} merges the hidden states specific to different tokenizations rather than the inputs and the preceding hidden states.
Formally, the transition equations are defined as follows:
\begin{align}
\label{Word-Lattice-GRU2-RTK-Func}
&\hspace{-0.1in} r_{t}^{(k)} = \sigma(W^{(r)}x_{t}^{(k)}+U^{(r)}h_{pre}^{(k)}),\\
&\hspace{-0.1in} z_{t}^{(k)} = \sigma(W^{(z)}x_{t}^{(k)}+U^{(z)}h_{pre}^{(k)}),\\
&\hspace{-0.1in} \tilde{h}_{t}^{(k)} = \phi(Wx_{t}^{(k)}+U(r_{tk}\odot h_{pre}^{(k)}),
\end{align}
\begin{align}
\label{Word-Lattice-GRU2-HTK-Func}
&\hspace{-0.1in} h_{t}^{(k)} = z_{t}^{(k)}\odot h_{pre}^{(k)} + (1-z_{t}^{(k)})\odot \tilde{h}_{t}^{(k)},\\
\label{Word-Lattice-GRU2-HT-Func}
&\hspace{-0.1in} h_{t} = g(h_{t}^{(1)},h_{t}^{(2)},...),
\end{align}
where Eqs. (\ref{Word-Lattice-GRU2-RTK-Func})-(\ref{Word-Lattice-GRU2-HTK-Func}) calculate the gating vectors and the hidden state vectors depending on different tokenizations,
and Eq. (\ref{Word-Lattice-GRU2-HT-Func}) is used to produce the final hidden state vector at the time step $t$.

For the composition function $g$(*) involved in the two word-lattice based GRUs, we explore the following two functions:
1) \textbf{Pooling Operation Function} which enables our encoders to automatically capture the most important part for the source sentence modeling,
and 2) \textbf{Gating Operation Function} which is able to automatically learn the weights of components in word-lattice based GRUs.
Taking $h_t$ as an example, we define it as follows
\begin{align}\label{Gated-Operation-Equation}
h_{t} = \sum\limits^K_{k=1}\frac{\sigma(h_{tk}U^{(g)}+b^{(g)})}{\sum\limits^K_{k=1}\sigma(h_{tk}U^{(g)}+b^{(g)})}h_{tk}
\end{align}
where $U^{(g)}$ and $b^{(g)}$ are the matrix and the bias term, respectively.

\section{Experiments}
\subsection{Setup}
We evaluated the proposed encoders on NIST Chinese-English translation tasks.
Our training data consists of 1.25M sentence pairs extracted from 
LDC2002E18, LDC2003E07, LDC2003E14, Hansards portion of LDC2004T07, LDC2004T08 and LDC2005T06,
with 27.9M Chinese words and 34.5M English words.
We chosed the NIST 2005 dataset as the validation set and the NIST 2002, 2003, 2004, 2006, and 2008 datasets as test sets.
The evaluation metric is case-insensitive BLEU \cite{Papineni:ACL02} as calculated by the \texttt{multi-bleu.perl} script.
To alleviate the impact of the instability of NMT training, we trained NMT systems five times for each experiment and reported the average BLEU scores.
Furthermore, we conducted paired bootstrap sampling \cite{Koehn:04} to test the significance in BLEU score differences.
To obtain lattices of Chinese sentences, we used the toolkit\footnote{http://nlp.stanford.edu/software/segmenter.html\#Download} released by Stanford to train word segmenters on \emph{CTB}, \emph{PKU}, and \emph{MSR} corpora.

To train neural networks efficiently, we used the most frequent 50K words in Chinese and English as our vocabularies.
Such vocabularies cover 98.5$\%$, 98.6$\%$, 99.3$\%$, and 97.3$\%$ Chinese words in \emph{CTB}, \emph{PKU}, \emph{MSR}, and lattice corpora, and 99.7$\%$ English words, respectively.
In addition, all the out-of-vocabulary words are mapped into a special token UNK.
We only kept the sentence pairs that are not longer than 70 source characters and 50 target words, which cover 89.9$\%$ of the parallel sentences.
We apply \emph{Rmsprop} \cite{Graves:Arxiv13} (momentum = 0, $\rho$ = 0.99, and $\epsilon$ = 1$\times$ $10^{-4}$)
to train models until there is no BLEU improvement for 5 epochs on the validation set.
During this procedure, we set the following hyper-parameters:
word embedding dimension as 320,
hidden layer size as 512,
learning rate as $5\times10^{-4}$,
batch size as 80,
gradient norm as 1.0.
All the other settings are the same as in \cite{Bahdanau:ICLR15}.

\subsection{Baselines}
We refer to the attention-based NMT system with the proposed encoders as \emph{\textbf{LatticeNMT}}, which has four variants with different network units and composition operations.
We compared them against the following state-of-the-art SMT and NMT systems:

\begin{itemize}
\item \emph{\textbf{Moses}}\footnote{http://www.statmt.org/moses/}: an open source phrase-based translation system with default configurations and a 4-gram language model trained on the target portion of training data.
\item \emph{\textbf{RNNSearch}} \cite{Bahdanau:ICLR15}: an in-house attention-based NMT system, which has slightly better performance and faster speed than \emph{GroundHog}.\footnote{https://github.com/lisa-groundhog/GroundHog}
    In addition to the \emph{RNNSearch} with word-based segmentations, we also compared to that with character-based segmentations to study whether word boundary information is helpful for NMT.
\item \emph{\textbf{MultiSourceNMT}}: an attention-based NMT system which also uses many tokenizations of each source sentence as input.
    However, it performs combination at the final time step, significantly different from the proposed word-lattice based GRUs.
    Similarly, it has two variants with different composition functions.
\end{itemize}

\subsection{Overall Performance}
\begin{table*}[htbp]
	\centering
	\small
	\begin{tabular}{l|c|c|p{25pt}|p{25pt}p{25pt}p{25pt}p{25pt}p{25pt}|p{25pt}}
		\textbf{System} & \textbf{Unit} & \textbf{Input} & \textbf{MT05} & \textbf{MT02} & \textbf{MT03} & \textbf{MT04} & \textbf{MT06} & \textbf{MT08} & \textbf{ALL}\\
        \hline
        \hline
		\multirow{4}{*}{\emph{Moses}} & & \emph{CTB} & 31.70 & 33.61 & 32.63 & 34.36 & 31.00 & 23.96 & 31.25 \\
		 	                                   & ------ & \emph{PKU} & 31.47 & 33.19 & 32.43 & 34.14 & 30.81 & 23.85 & 31.04 \\
		 	                                   & & \emph{MSR} & 31.08 & 32.44 & 32.08 & 33.78 & 30.28 & 23.60 & 30.56 \\
                                               \cline{3-10}
                                               & & \emph{Lattice} & 31.96 & 33.44 & 32.54 & 34.61 & 31.36 & 24.13 & 31.50 \\
                                        	   \hline
		\multirow{4}{*}{\emph{RNNSearch}}  & \multirow{4}{*}{\emph{GRU}} & \emph{Char} & 30.30 & 33.51 & 31.98 & 34.34 & 30.50 & 22.65 & 30.74 \\
                                                    \cline{3-10}
                                                    & & \emph{CTB} & 31.38 & 34.95 & 32.85 & 35.44 & 31.75 & 23.33 & 31.78 \\
        	                                        & & \emph{PKU} & 31.42 & 34.68 & 33.08 & 35.32 & 31.61 & 23.58 & 31.76 \\
                                                    & & \emph{MSR} & 29.92 & 34.49 & 32.06 & 35.10 & 31.23 & 23.12 & 31.35 \\
		                                            \hline
        \multirow{2}{*}{\emph{MultiSourceNMT}} & \emph{GRU} + \emph{Pooling} & \multirow{2}{*}{\emph{CTB+PKU+MSR}} & 30.27 & 34.73 & 31.76 & 34.79 & 31.36 & 23.54 & 31.34 \\
                                               & \emph{GRU} + \emph{Gating} &  & 31.95 & 35.14 & 33.36 & 35.84 & 32.09 & 23.60 & 32.13 \\
		\hline
        \multirow{4}{*}{\emph{LatticeNMT}} & \emph{SWL-GRU} + \emph{Pooling} & \multirow{4}{*}{\emph{Lattice}} & 32.27 & 35.35 & 34.20 & 36.49 & 32.27 & 24.52 & 32.69$^{\dag,\ddag}$\\
                                          & \emph{SWL-GRU} + \emph{Gating} &  & 32.07 & \textbf{35.94} & 34.01 & 36.48 & 32.51 & 24.44 & 32.79$^{\dag,\ddag,\uparrow}$ \\
                                          & \emph{DWL-GRU} + \emph{Pooling} &  & 32.18 & 35.53 & 33.94 & 36.42 & 32.61 & 24.34 & 32.71$^{\dag,\ddag,\uparrow}$ \\
                                          & \emph{DWL-GRU} + \emph{Gating} &  & \textbf{32.40} & 35.75 & \textbf{34.32} & \textbf{36.50} & \textbf{32.77} & \textbf{24.84} & \textbf{32.95}$^{\dag,\ddag,\uparrow}$\\

	\end{tabular}
	\caption{\label{Experiment-Result1}
Evaluation of translation quality.
$\dag$, $\ddag$ and $\uparrow$ indicate statistically significantly better than (\emph{p}$<$0.01) the best results of \emph{Moses}, \emph{RNNSearch} and \emph{MultiSourceNMT} system, respectively.
We highlight the highest BLEU score in bold for each set.
}
\end{table*}
\begin{table*}[htbp]
	\centering
	\small
	\begin{tabular}{l|c|c|p{25pt}p{25pt}p{25pt}p{25pt}p{25pt}|p{25pt}}
		\textbf{System} & \textbf{Unit} &\textbf{Input} & \textbf{MT02} & \textbf{MT03} & \textbf{MT04} & \textbf{MT06} & \textbf{MT08} & \textbf{ALL}\\
        \hline
        \hline
		\multirow{4}{*}{\emph{LatticeNMT}} & \multirow{4}{*}{\emph{DWL-GRU} + \emph{Gating}} & \emph{Lattice} & 35.75 & 34.32 & 36.50 & 32.77 & 24.84 & 32.95  \\
                                                       & & \emph{CTB} & 33.31 & 31.44 & 34.05 & 30.09 & 22.59 & 29.97  \\
                                                       & & \emph{PKU} & 32.58 & 31.22 & 33.62 & 29.72 & 22.68 & 29.68  \\
                                                       & & \emph{MSR} & 30.53 & 28.82 & 32.37 & 28.08 & 21.83 & 28.14  \\
	\end{tabular}
	\caption{\label{Experiment-Result3}
Experiment results of \emph{LatticeNMT} decoding the test sets with 1-best segmentations.
}
\end{table*}
Table \ref{Experiment-Result1} reports the experiment results.
Obviously, \emph{LatticeNMT} significantly outperforms non-lattice NMT systems in all cases.
When using \emph{DWL-GRU} with the \emph{gating} composition operation, the \emph{LatticeNMT} system outperforms \emph{Moses}, \emph{RNNSearch}, \emph{MultiSourceNMT} by at least gains of 1.45, 1.17, and 0.82 BLEU points, respectively.

\noindent\textbf{Parameters}.
Word- and character-based \emph{RNNSearch} systems have 55.5M and 44.2M parameters, respectively.
Only NMT systems with gating operation,
introduce no more than 2.7K parameters over those parameters of \emph{RNNSearch}.

\noindent\textbf{Speed}.
We used a single GPU device TitanX to train models.
It takes one hour to train 9,000 and 4,200 mini-batches for word- and character-based \emph{RNNSearch} systems, respectively.
The training speeds of \emph{MultiSoucceNMT} and \emph{LatticeNMT} systems are slower than that of word-based \emph{RNNSearch}: about 4,800$\sim$6,000 mini-batches are processed in one hour.

From the table, we further have the following findings:

(1) On all word-segmented corpora (e.g., \emph{CTB}, \emph{PKU}, \emph{MSR}), \emph{RNNSearch} performs better than character-based NMT system, which demonstrates that for Chinese which is not morphologically rich, word boundary information is very useful for source sentence encoding in NMT.
For this, we speculate that the character-level encoding has two disadvantages in comparison with the word-level encoding.
First, when a source sentence is represented as a sequence of characters rather than words, the sequence length grows significantly.
However, translating long sequences still remains a great challenge for the attention-based NMT \cite{Bentivogli:16}.
Second, the character-level encoder is unable to capture word-level interactions for learning the representations of entire sentences.

(2) Multiple inputs with different word segmentation standards are useful for NMT.
Furthermore, multiple inputs based on word lattices achieve better results.
The underlying reason may be that rather than only at final time step, all hidden states of word-lattice based encoders at potential word boundaries are influenced by many different tokenizations.
This enables different tokenizations to fully play complementary roles in learning source sentence representations.

(3) No matter which composition function is used, \emph{DWL-GRU} is slightly more effective than \emph{SWL-GRU}.
These results accord with our intuition since \emph{DWL-GRU} exploits multiple tokenizations at a more fine-grained level than \emph{SWL-GRU}.

\subsection{Analysis}
In order to take a deep look into the reasons why our encoders are more effective than the conventional encoder,
we study the 1-best translations produced by two systems.
The first system is the \emph{RNNSearch} using \emph{CTB} segmentations, denoted by \emph{\textbf{RNNSearch}}(\emph{\textbf{CTB}}).
It yields the best performance among all non-lattice NMT systems.
The other is \emph{\textbf{LatticeNMT}}(\emph{\textbf{DG}}) using \emph{DWL-GRU} with the \emph{gating} composition operation, which performs better than other \emph{LatticeNMT} systems.
We find that the utilization of word lattice brings the following advantages:

\begin{figure*}[!t]
\centering
\includegraphics[height=3.5cm,width=13cm]{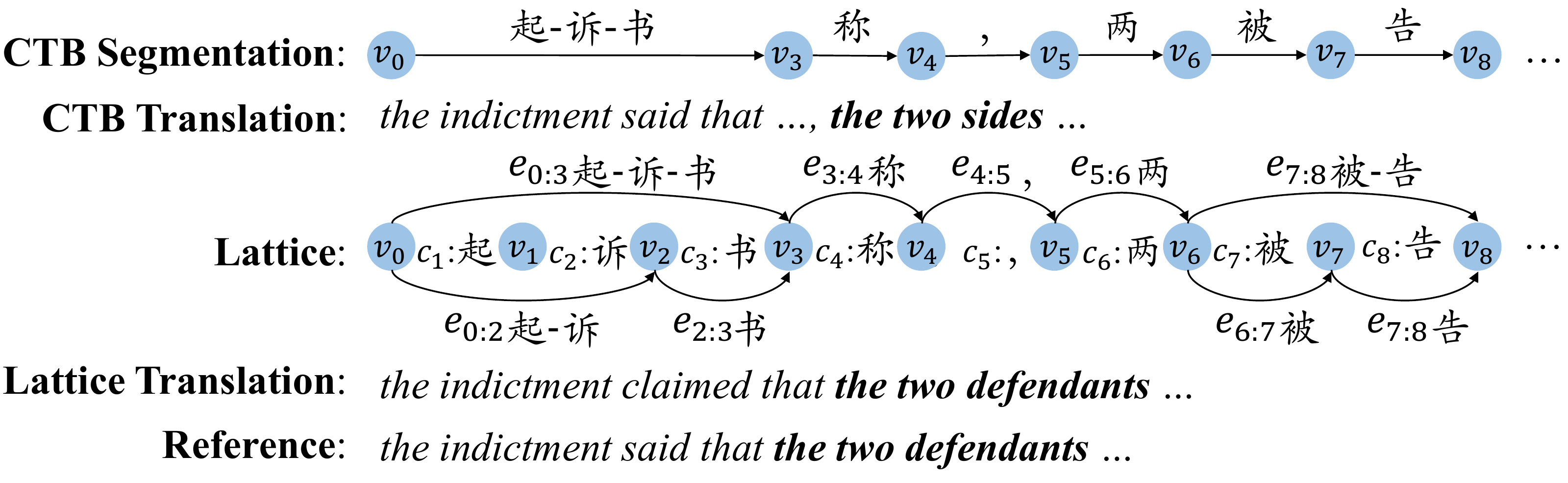}
\caption{
\label{CTBLatticeTran}
Translations of \emph{RNNSearch}(\emph{CTB}) and \emph{LatticeNMT}(\emph{DG}).
}
\end{figure*}

(1) Word lattices alleviate 1-best tokenization errors that further cause translation errors.
\begin{table}[htbp]
	\centering
	\small
	\begin{tabular}{l|cccc}
		 & \textbf{CTB} & \textbf{PKU} & \textbf{MSR} & \textbf{Lattice}\\
        \hline
        \hline
		Percentage & 93.53\% & 94.20\% & 96.36\% & \textbf{97.14\%} \\
	\end{tabular}
	\caption{\label{Experiment-Result3}
The percentages of character spans covered by in-vocabulary words.
}
\end{table}
As illustrated in Figure \ref{CTBLatticeTran},
in the source sentence ``$c_1$起$c_2$诉$c_3$书$c_4$称$c_5$，$c_6$两$c_7$被$c_8$告...'',
the character subsequence $c_{7:8}$``被 \ 告'' is incorrectly segmented into two words ``被'' and ``告'' by the \emph{CTB} segmenter.
As a result, the word sequence ``两 \ 被 \ 告'' is incorrectly translated into ``\emph{the two sides}''.
In contrast, \emph{LatticeNMT}(\emph{DG}) automatically chooses the right tokenization ``被-告'' due to its higher bidirectional gating weights (0.982 and 0.962).
This allows the NMT system to choose the right translation ``\emph{the two defendants}'' for the word sequence ``两 \ 被-告''.

(2) Word lattices endow NMT encoders with highly expressible and flexible capabilities to learn the semantic representations of input sentences.
To test this, we considered 1-best segmentations as a special kind of word lattice and used the trained \emph{LatticeNMT}(\emph{DG}) system to decode the test sets with 1-best \emph{CTB}, \emph{PKU}, \emph{MSR} segmentations, respectively.
Intuitively, if \emph{LatticeNMT}(\emph{DG}) mainly depends on 1-best segmentations, it will have similar performances on lattice and the 1-best segmentation corpora.
From Table \ref{Experiment-Result3}, we find that when decoding paths are constrained by 1-best segmentations, the performance of \emph{LatticeNMT}(\emph{DG}) system degrades significantly.
Therefore, our \emph{LatticeNMT} system is able to explore tokenizations with different annotation standards.

(3) We also find that the lattice-based method reduces the number of UNK words to a certain extent.
We calculated the percentages of the maximal character spans of inputs covered by in-vocabulary words, and compared the \emph{CTB}, \emph{PKU}, \emph{MSR} and lattice corpora.
Results are shown in Table \ref{Experiment-Result3}.
We observe that the lattice corpus has the highest percentage of character spans covered by in-vocabulary words.

\section{Related Work}
Our work is related to previous studies that learn sentence representations with deep neural networks.
Among them, the most straightforward method is the neural \emph{Bag-of-Words} model, which, however, neglects the important information of word order.
Therefore, many researchers resort to order-sensitive models, falling into one of two classes:
(1) \emph{sequence models} and (2) \emph{topological models}.
The most typical sequence models are RNNs 
with LSTM \cite{Hochreiter:NC97,Sutskever:NIPS14,Zhu:15,Liu:15} or GRU \cite{Cho:EMNLP14}.
Further, some researchers extend standard RNNs to non-sequential ones,
such as \emph{multi-dimensional RNNs} \cite{Graves:07}, \emph{grid RNNs} \cite{Kalchbrenner:15} and \emph{higher order RNNs} \cite{Soltani:ARXIV16}.
In topological models, sentence representations are composed following given topological structures over words
\cite{Socher:ICML11,Hermann:13,Iyyer:14,Mou:15,Tai:ACL15,Le:15b}.
In addition to the aforementioned models, convolutional neural networks are also widely used to model sentences \cite{Collobert:11,Kalchbrenner:14,Kim:14,Hu:14}.

Concerning  NMT, the conventional NMT relies almost exclusively on word-level source sentence modelling with explicit tokenizations \cite{Sutskever:NIPS14,Cho:EMNLP14,Bahdanau:ICLR15}, which tends to suffer from the problem of unknown words.
To address this problem, researchers have proposed alternative character-based modelling.
In this respect, Costa-Juss$\grave{a}$ and Fonollosa \shortcite{Marta:ACL16} applied character-based embeddings in combination with convolutional and highway layers 
to produce word embeddings.
Similarly, Ling et al. \shortcite{Wang:Arxiv15} used a bidirectional LSTM to generate semantic representations of words based on character embeddings.
A slightly different approach was proposed by Lee et al. \shortcite{Lee:WAT15}, where they explicitly marked each character with its relative location in a word.
Recently, Chung et al. \shortcite{Chung:16} evaluated a character-level decoder without explicit segmentations for NMT.
However, their encoder is still a subword-level one.
Overall, word boundary information is very important for the encoder modelling of NMT.

In contrast to 
the above approaches,
we incorporate word lattice into RNN encoders,
which, to the best of our knowledge, has never been investigated before in NMT.
The most related models to ours are those proposed by Tai et al., \shortcite{Tai:ACL15}, Le et al., \shortcite{Le:15b}, Kalchbrenner et al., \shortcite{Kalchbrenner:15}, Soltani and Jiang \shortcite{Soltani:ARXIV16}, Ladhak et al., \shortcite{Ladhak:INTERSPEECH16}.
Tai et al., \shortcite{Tai:ACL15} presented \emph{tree-structured LSTMs},
while Le and Zuidema \shortcite{Le:15b} further introduced the forest convolutional network for sentence modelling.
Kalchbrenner et al., \shortcite{Kalchbrenner:15} studied \emph{grid RNNs} where the inputs are arranged in a multi-dimensional grid.
In \emph{higher order RNNs} proposed by Soltani and Jiang \shortcite{Soltani:ARXIV16}, more memory units are used to record more preceding states, which are all recurrently fed to the hidden layers as feedbacks through different weighted paths.
Our work is also significantly different from these models.
We introduce word lattices rather than parse trees and forests to improve sentence modelling,
and thus our network structures depend on the input word lattices, significantly different from the prefixed structures of \emph{grid RNNs} or \emph{high order RNNs}.
More importantly, our encoders are able to simultaneously process multiple input vectors specific to different tokenizations, while \emph{tree-structured LSTMs}, \emph{grid RNNs} and \emph{high order RNNs} deal with only one input vector at each time step.
Concurrent with our work, Ladhak et al. \shortcite{Ladhak:INTERSPEECH16} also presented a method to generalize RNNs to encode word lattices, and demonstrated its effectiveness on automatic speech recognition.

\section{Conclusions and Future Work}
This paper has presented word-lattice based RNN encoders for NMT.
Compared with the standard RNN encoders, our encoders simultaneously exploit the inputs and preceding hidden states depending on multiple tokenizations for source sentence modelling.
Thus, they not only alleviate error propagations of 1-best tokenizations, but also are more expressive and flexible than the standard encoder.
Experiment results on Chinese-English translation show that our encoders lead to significant improvements over a variety of baselines.

In the future, we plan to continue our work in the following directions.
In this work, our network structures depend on the word lattices of source sentences.
We will extend our models to incorporate the segmentation model into source sentence representation learning.
In this way, tokenization and translation are allowed to collaborate with each other.
Additionally, we are interested in exploring better combination strategies to improve our encoders.


\bibliographystyle{aaai}
\bibliography{LatticeRNN_NMT}

\end{CJK*}

\end{document}